\documentclass[11pt,letterpaper]{article}
\usepackage[T1]{fontenc}
\usepackage{times}
\usepackage[margin=1in,headheight=14pt]{geometry}
\usepackage{amsmath,amssymb,bm}
\usepackage{graphicx,booktabs,array}
\usepackage{microtype}
\usepackage{xcolor}
\usepackage{xurl}
\usepackage[round,authoryear]{natbib}
\usepackage[hidelinks,pdfversion=1.7]{hyperref}
\usepackage{fancyhdr}
\usepackage[font=small,labelfont=bf]{caption}

\graphicspath{{figures/}}
\DeclareRobustCommand{\remory}{\textsc{Remory}}
\DeclareRobustCommand{\compactcmd}{\texttt{/compact}}
\newcommand{\doi}[1]{doi: \href{https://doi.org/#1}{\nolinkurl{#1}}}
\hypersetup{pdftitle={Remory: Learning Residual Memory for Context Compaction},
    pdfauthor={},pdfsubject={Preprint}}
\setcitestyle{citesep={;},aysep={,},yysep={;}}
\renewcommand{\headrulewidth}{0pt}

\title{\LARGE\textsc{Remory: Learning Residual Memory\\for Context Compaction}}
\author{%
    Hanchen Xia\textsuperscript{1}\quad
    Baoyou Chen\textsuperscript{1,3}\quad
    Yutang Ge\textsuperscript{4}\quad
    Naihao Deng\textsuperscript{5}\\[3pt]
    Senqiao Yang\textsuperscript{6}\quad
    Zilong Dong\textsuperscript{7}\quad
    Weihao Yuan\textsuperscript{8}\quad
    Siyu Zhu\textsuperscript{3,1,2,*}\\[7pt]
    {\small\textsuperscript{1}Shanghai Academy of AI for Science\quad
        \textsuperscript{2}Shanghai Innovation Institute}\\
    {\small\textsuperscript{3}Fudan University\quad
        \textsuperscript{4}Shanghai Jiao Tong University}\\
    {\small\textsuperscript{5}University of Michigan\quad
        \textsuperscript{6}The Chinese University of Hong Kong}\\
    {\small\textsuperscript{7}Alibaba Group\quad
        \textsuperscript{8}Nanjing University}\\[5pt]
    {\small\texttt{\{xiahanchen, chenbaoyou\}@sais.org.cn}}\\[3pt]
    {\small\textsuperscript{*}Corresponding author.}}

\hypersetup{pdfauthor={Hanchen Xia, Baoyou Chen, Yutang Ge, Naihao Deng,
    Senqiao Yang, Zilong Dong, Weihao Yuan, Siyu Zhu}}

\date{}
\makeatletter
\renewcommand{\section}{\@startsection{section}{1}{\z@}%
    {-2.5ex plus -0.5ex minus -0.2ex}{1.2ex plus 0.2ex}%
    {\normalfont\Large\bfseries}}
\renewcommand{\subsection}{\@startsection{subsection}{2}{\z@}%
    {-2ex plus -0.5ex minus -0.2ex}{0.8ex plus 0.2ex}%
    {\normalfont\large\bfseries}}
\renewcommand{\maketitle}{%
    \begin{center}
        {\@title\par}
        \vspace{12pt}
        {\normalsize\@author\par}
    \end{center}
    \vspace{5pt}}
\makeatother

\begin{document}
\maketitle
\thispagestyle{plain}

\begin{abstract}
Long-horizon agents compact their history to continue within a finite
context window, but a textual summary alone may not support every
subsequent decision. We introduce \remory{}, a neural memory network
that supplements the summary with a bounded sequence of soft memory
tokens. Given the history and summary, the network learns to generate
tokens that help a frozen LLM approximate the continuation it would
produce with the full history. The tokens are conditioned on the summary
and appended after it, forming an analogue of a residual connection along
the sequence dimension. On SummHay, \remory{} improves source attribution
at nearly unchanged insight coverage and approaches the full-context
joint score using only 5.2\% of the input positions. Across long-horizon
agent benchmarks, Qwen3.8-27B and GLM-5.3-Flash show consistent gains
with residual memory. Both models also exhibit substantially fewer
repeated tool outputs and tool errors on BrowseComp and Terminal-Bench 2.1.
\end{abstract}

\section{Introduction}
\label{sec:introduction}

\begin{figure}[t]
    \centering
    \includegraphics[width=\linewidth]{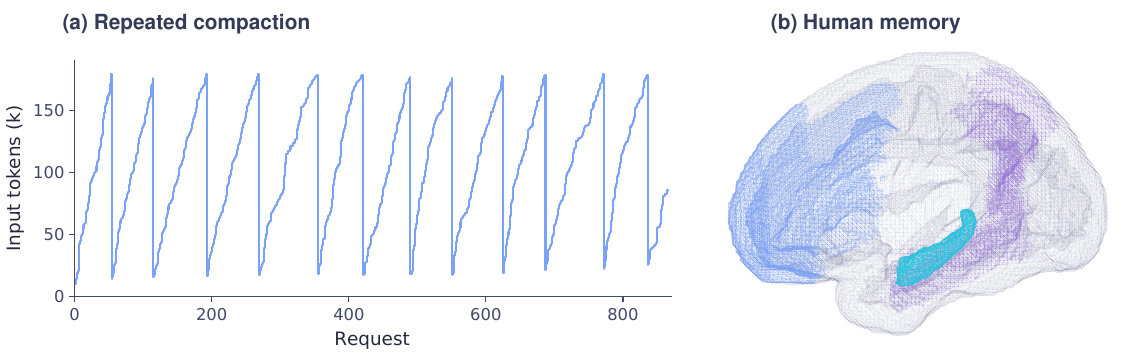}
    \caption{Context compaction and human memory.
    (a) Input length across successive requests in an agent run, showing
    repeated growth and sharp reductions.
    (b) Memory systems illustrated using the Harvard--Oxford
    atlas~\citep{harvardoxfordatlas}: prefrontal regions (blue), hippocampus
    (cyan), and association cortex (purple).}
    \label{fig:context-and-memory}
\end{figure}

Long-horizon agents work within a finite context window. As observations,
tool outputs, and decisions accumulate, they must compact their history
to continue. A common practice today is to replace earlier interaction
with a textual summary. For example, popular coding agents such as
Claude Code and Codex CLI provide \compactcmd{} to free context
space~\citep{anthropic2026context,openai2026codexcommands}.
The summary becomes a checkpoint from which the agent resumes its work.
Input length follows a sawtooth pattern: it grows with interaction,
drops at compaction, and grows again (Figure~\ref{fig:context-and-memory}a).

A textual checkpoint changes how prior context is available to the agent.
Preserving the main thread does not guarantee that the summary will
support every subsequent decision, and the relevance of earlier context
can change as the task unfolds. We do not assume that the information
needed beyond the summary can be identified or enumerated in advance.
The challenge is to sustain useful access to prior context across
compaction without keeping the entire interaction in the active window.

\begin{figure}[t]
    \centering
    \includegraphics[width=\linewidth]{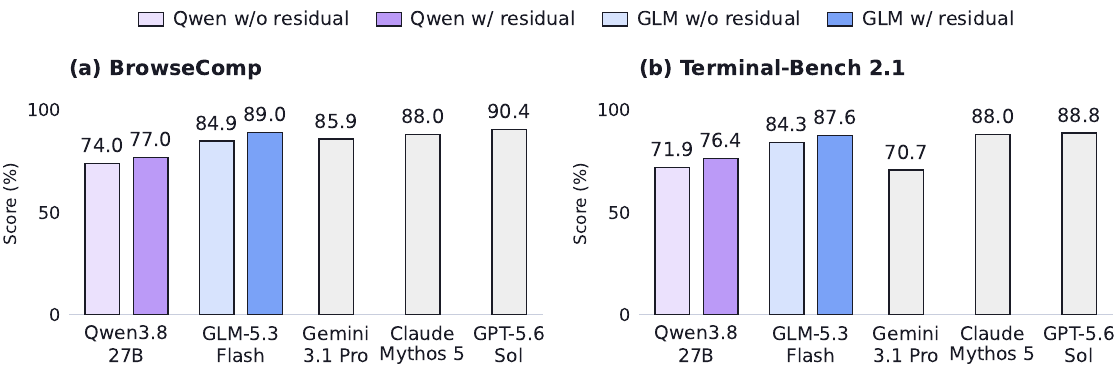}
    \caption{BrowseComp and Terminal-Bench 2.1 scores. Pale and solid bars
    compare the same local actor without and with residual memory.
    Gray bars are published frontier references, with their original
    harnesses and protocols~\citep{google2026gemini31pro,terminalbench2026v21,openai2026gpt56};
    they are not controlled comparisons with our runs.}
    \label{fig:headline-results}
\end{figure}

Human cognition illustrates how a limited active workspace can support
sustained work. While contemporary agent contexts span roughly
0.2--1.0 million token positions, human working memory holds only a few
chunks under controlled conditions~\citep{baddeley2012working,cowan2001magical}.
Yet people can pursue complex projects over months or years, repeatedly
drawing on details, experience, and learned skills in long-term
memory~\citep{ericsson1995longterm,squire2015conscious}. This ability depends
on access to stored knowledge as well as active maintenance.
Hippocampal indexing theory describes one form of \emph{neural indexing}:
partial cues activate an index that reinstates distributed cortical
representations of an earlier experience~\citep{teyler2007indexing}.
Figure~\ref{fig:context-and-memory}b illustrates the complementary memory
systems involved~\citep{goldmanrakic1995cellular,kumaran2016learning}.
For agents, this suggests pairing an explicit checkpoint with a learned
neural representation that keeps prior context useful as work continues.

We introduce \remory{}, a neural memory network for context compaction.
Given a history and its generated summary, we train the network to
generate a bounded sequence of soft memory tokens that can be appended
after the summary, helping a frozen LLM approximate the continuation
it would produce with the full history. We call this \emph{residual memory}
because the soft tokens are both conditioned on the summary and appended
after it, forming an analogue of a residual connection along the sequence
dimension. The summary supplies a readable checkpoint, while the learned
tokens supplement it during subsequent decisions. Exact records remain
accessible through ordinary archive and file-retrieval tools.

On SummHay, residual tokens improve source attribution at nearly unchanged
insight coverage. This supports an associative interpretation: the memory
helps connect recovered claims to supporting evidence. Across long-horizon
agent benchmarks, Qwen3.8-27B and GLM-5.3-Flash show consistent gains
with residual memory. Residual memory also reduces repeated tool outputs
and tool errors for both models on BrowseComp and Terminal-Bench 2.1.
\textsc{Remory} pushes the 320B-parameter
GLM-5.3-Flash~\citep{zai2026glm53flash} towards frontier-level performance
(Figure~\ref{fig:headline-results}).

\section{Residual Memory}
\label{sec:method}

\begin{figure}[!htbp]
    \centering
    \includegraphics[width=\linewidth]{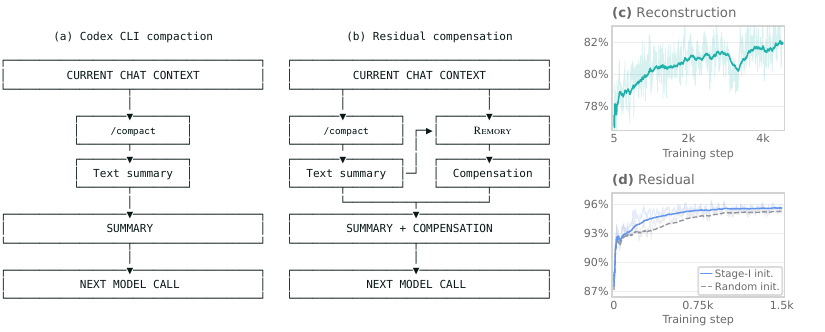}
    \caption{Residual compensation and training.
    (a)~Codex CLI passes the \compactcmd{} summary to the next actor call.
    (b)~The \remory{} network uses that summary to condition memory construction from the
    active context; summary and continuous memory jointly support subsequent actions.
    Archive access remains an ordinary tool action in either setting.
    (c,d)~Qwen3.8-27B teacher--student top-1 agreement during Stage I
    reconstruction and Stage II residual training. The residual panel compares
    reconstruction initialization with
    random initialization under the same residual objective and data order.
    Curves are optimization diagnostics from the training streams.}
    \label{fig:context-management}
    \label{fig:two-stage-training}
\end{figure}

\subsection{A memory supplement to the summary}

Residual memory is constructed by conditioning on a generated summary
and appending learned soft tokens after it
(Figure~\ref{fig:context-management}(a,b)).
At compaction boundary $i$,
$H_i$ contains the previous summary and residual memory, when present,
together with the interaction since the last boundary. Given the new
summary $S_i$, the memory network constructs
\begin{equation}
    M_i=\mathcal{C}_{\phi}(H_i;S_i),\qquad |M_i|\leq m.
    \label{eq:memory}
\end{equation}
$M_i$ is a sequence of vectors in the frozen actor's input space.
The concatenation $[S_i;M_i]$ forms an analogue of a residual connection
along the sequence dimension. The actor reads persistent instructions, this
sequence, and then new interaction. The same process repeats at later compactions,
allowing memory to carry information across successive boundaries.
Archive excerpts enter through ordinary tool results.

The network uses frozen actor features and learned queries to predict
memory vectors in parallel, with gated cross-attention to the summary.
Each block of up to 1,024 source positions produces 64 soft tokens.
Hierarchical compression bounds the final memory at 4,096 slots while
preserving source order. Source encoding and intermediate representations
may be larger.

For Qwen3.8-27B, we adapt the DFlash
architecture~\citep{chen2026dflash}; for GLM-5.3-Flash, we use DFlash-2
with grouped dynamic convolutions~\citep{inco2026dflash2}.
The resulting memory networks have 1.94B and 1.502B trainable parameters,
respectively. Their output is continuous memory rather than vocabulary
tokens. Gradients through the frozen actor train this interface without
changing the actor's parameters.

\subsection{Learning through a frozen actor}
\label{sec:training}

Figure~\ref{fig:two-stage-training}(c,d) shows optimization behavior
during the two training stages.

\textbf{Stage I: reconstruction.} The teacher receives a source passage
and an instruction to repeat it. The student replaces the passage with
its soft memory and receives the same instruction. Both predict the
source text, teaching the memory network to preserve source information.
We use text fields from MMFineReason-Full-2.3M~\citep{lin2026mmfinereason}.

\textbf{Stage II: summary-conditioned continuation.} Starting from Stage I,
the teacher receives the original interaction history. The student receives
the corresponding fixed summary and memory constructed from that history.
Both predict the same subsequent assistant response. The memory encoder
never receives the future response. This stage uses reasoning and coding
trajectories, including prompts from
Open-PerfectBlend~\citep{labonne2024openperfectblend}, with precomputed
summaries and Qwen-generated responses. Examples follow a curriculum
ordered by preceding-context length.

In both stages, the teacher and student share the same frozen actor
$p_\theta$. For a target response $y$, their predictions at position $j$ are
\begin{equation}
    p_j=p_\theta(\cdot\mid C_{\mathrm{full}},y_{<j}),\qquad
    q_j=p_\theta(\cdot\mid C_{\mathrm{mem}},y_{<j}),
\end{equation}
where both contexts include the persistent instructions. We optimize
the memory parameters through student-to-teacher reverse KL:
\begin{equation}
    \mathcal{L}(\phi)=
    \frac{\sum_j w_j D_{\mathrm{KL}}(q_j\Vert p_j)}{\sum_j w_j}.
    \label{eq:objective}
\end{equation}
The loss covers the full vocabulary and complete response, with positive
weights that mildly emphasize earlier positions. Only $\phi$ is updated.
In Stage II, the summary is already available, directing memory toward
the remaining prediction gap. Memory may overlap with summary content;
it need not reconstruct every omitted fact.

\section{Experiments}
\label{sec:experiments}

\subsection{Source attribution after compaction}
\label{sec:summhay}

SummHay evaluates query-relevant insight recovery and source attribution
from long document collections~\citep{laban2024summhay}. We evaluate all
92 queries from ten collections with Qwen3.8-27B under four representations:
raw context, summary alone, summary plus an additional text supplement
(summary++), and summary plus residual memory. Summary++ uses the same
additional-position budget as the memory condition. The actor, query,
decoding, and base summary are fixed across the compacted conditions.
Answer generation has no tools, archive access, or source retrieval;
gold insights and citations are used only for scoring.

\begin{table}[t]
    \centering
    \caption{SummHay on the same 92 queries. Summary++ adds text under
    the residual memory's additional-position budget. Higher is better;
    all queries, including generation and formatting failures, are retained.}
    \label{tab:summhay}
    \small
    \setlength{\tabcolsep}{10pt}
    \begin{tabular}{@{}lrrr@{}}
        \toprule
        Representation & Coverage & Citation F1 & Joint score \\
        \midrule
        Raw context & 72.46 & 56.60 & 43.41 \\
        Summary & 67.55 & 56.98 & 39.84 \\
        Summary++ & 67.52 & 56.94 & 39.78 \\
        Summary + \remory{} & 67.95 & \textbf{61.03} & 43.39 \\
        \bottomrule
    \end{tabular}
\end{table}

Coverage measures recovered insights; citation F1 measures attribution;
the joint score combines both. We use GPT-5.6-Sol with the benchmark
coverage prompts. Coverage and joint score average over all 621 reference
insights, while citation F1 averages over covered insights.
Adding residual memory increases citation F1 by 4.04 points and joint
score by 3.55, with a coverage change of 0.40
(Table~\ref{tab:summhay}; gains are computed before rounding).
A paired collection bootstrap gives a 95\% interval of $[+0.74,+6.21]$
for the joint-score gain, conditional on the generated and judged outputs.
On the 553 insights covered by both summary conditions, citation F1 still
rises from 57.31 to 61.41. The larger attribution gain at nearly unchanged
coverage suggests that memory helps connect recovered claims to evidence.

Mean inputs contain 96,872 positions for raw context, 1,393 for summary
alone, and 5,024 for summary plus memory. The latter uses 5.2\% of the
raw-context positions and approaches its joint score, with lower coverage
and higher citation F1. The equal-budget text supplement remains close
to summary alone. In this setting, the learned supplement preserves
attribution cues more effectively than allocating the same extra
positions to generated text.

\subsection{End-to-end agent performance}
\label{sec:agents}

We evaluate BrowseComp research tasks~\citep{wei2025browsecomp},
Terminal-Bench 2.1 command-line tasks~\citep{merrill2026terminalbench},
AutomationBench workflows~\citep{shepard2026automationbench}, and JobBench
professional tasks~\citep{li2026jobbench}. Each within-actor comparison
holds weights, the Codex CLI harness, prompts, tools, decoding, budgets,
and compaction policy fixed; both configurations can retrieve exact history.
Qwen and GLM use 262,144-position and one-million-position context windows,
respectively. The intervention adds up to 4,096 residual slots at compaction.

We use the full task sets: 1,266 BrowseComp and 89 Terminal-Bench tasks
for both actors (one run), and 600 AutomationBench and 65 JobBench tasks
for Qwen (five runs). Terminal-Bench and AutomationBench use task
verifiers; BrowseComp and JobBench use matched GPT-5.6-Sol judges.
Residual checkpoints complete both training stages.

\begin{table}[t]
    \centering
    \caption{Qwen3.8-27B workflow results: mean $\pm$ standard deviation
    over five runs. Costs are mean API-equivalent USD per task.}
    \label{tab:workflows}
    \small
    \setlength{\tabcolsep}{8pt}
    \begin{tabular}{@{}lrr@{\hspace{20pt}}rr@{}}
        \toprule
        & \multicolumn{2}{c}{AutomationBench} & \multicolumn{2}{c}{JobBench} \\
        \cmidrule(lr){2-3}\cmidrule(l){4-5}
        Configuration & Score & Cost & Score & Cost \\
        \midrule
        w/o residual & 35.5 $\pm$ 1.3 & \$0.59 & 33.4 $\pm$ 1.25 & \$3.09 \\
        w/ residual & \textbf{45.3} $\pm$ 1.0 & \$0.53 & \textbf{41.0} $\pm$ 1.41 & \$3.21 \\
        \bottomrule
    \end{tabular}
\end{table}

\begin{table}[t]
    \centering
    \caption{Terminal-Bench 2.1 and BrowseComp: one run per configuration.
    Costs are mean API-equivalent USD per task. Repeats and errors are
    tool-output counts over the full task set.}
    \label{tab:agents}
    \small
    \setlength{\tabcolsep}{3pt}
    \begin{tabular}{@{}llrrrr@{\hspace{9pt}}rrrr@{}}
        \toprule
        & & \multicolumn{4}{c}{Terminal-Bench 2.1} & \multicolumn{4}{c}{BrowseComp} \\
        \cmidrule(lr){3-6}\cmidrule(l){7-10}
        Actor & Residual & Score & Cost & Repeats & Errors & Score & Cost & Repeats & Errors \\
        \midrule
        Qwen3.8-27B & w/o & 71.9 & \$4.29 & 459 & 933 & 74.0 & \$5.39 & 25,801 & 25,244 \\
        & w/ & \textbf{76.4} & \$2.86 & 377 & 568 & \textbf{77.0} & \$4.69 & 18,105 & 12,698 \\
        GLM-5.3-Flash & w/o & 84.3 & \$3.27 & 334 & 664 & 84.9 & \$5.65 & 8,316 & 7,498 \\
        & w/ & \textbf{87.6} & \$3.05 & 240 & 496 & \textbf{89.0} & \$5.00 & 5,834 & 5,412 \\
        \bottomrule
    \end{tabular}
\end{table}

Qwen improves by 9.8 points on AutomationBench and 7.6 on JobBench
(Table~\ref{tab:workflows}), extending the gains to business workflows
and professional-task rubric credit. It gains 4.5 points on Terminal-Bench
and 3.0 on BrowseComp; GLM gains 3.3 and 4.1, respectively
(Table~\ref{tab:agents}). The improvements therefore also hold for a
stronger actor with a larger context window. Published frontier scores
in Figure~\ref{fig:headline-results} provide context; the controlled
evidence comes from the within-actor comparisons.

Repeated tool outputs fall by 17.9--29.8\% and errors by 25.3--49.7\%
across the four actor--benchmark pairs in Table~\ref{tab:agents}.
Repeats count identical outputs from the same tool within a task, excluding
planning and polling; errors count failed commands or explicit tool errors.
These aggregate reductions are consistent with better use of earlier
feedback, although complete trajectories differ in their actions and length.

Generation cost falls in every reported comparison except JobBench, where
the higher score accompanies a rise from \$3.09 to \$3.21 per task.
The estimates price actor and summary tokens at September 2026 uncached provider rates,
counting residual positions as input~\citep{alibaba2026modelpricing,zai2026pricing}.
They exclude source-feature encoding, memory-network execution, and tools,
so they measure token-priced generation rather than total execution cost.

\subsection{The first action after compaction}

We examine 488 paired next actions from 100 sampled Qwen BrowseComp
trajectories, holding the persistent prompt and summary fixed.
Following rubric-based planning evaluation~\citep{yu2026webanchor},
GPT-5.6-Sol compares anonymized actions against criteria fixed from
the preceding history. Candidate reasoning is withheld and counterfactual
calls remain unexecuted.

Residual actions receive 345 wins (70.7\%), 21 ties (4.3\%), and 122 losses
(25.0\%). With the checkpoint fixed, the preference links the memory
channel to the first post-compaction decision. It is a judgment of proposed
actions, not a measurement of the outcomes those counterfactual actions
would have produced.

\subsection{Case studies}
\label{sec:cases}

\begin{figure}[t]
    \centering
    \includegraphics[width=\linewidth]{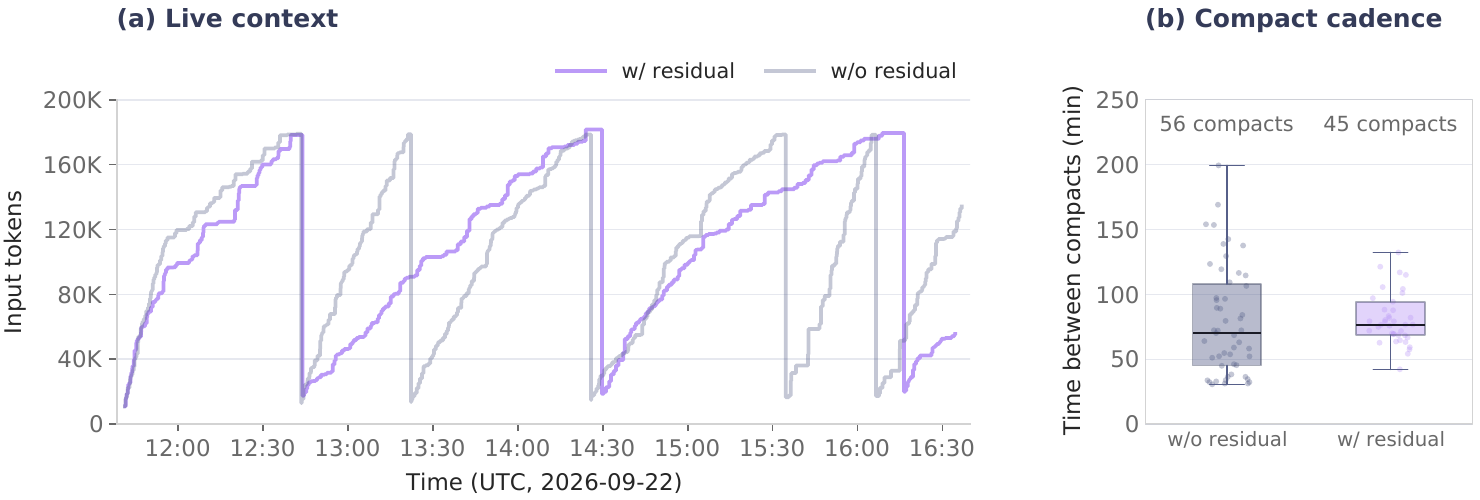}
    \caption{Live Trading context use. (a) An illustrative September 22
    trace. (b) Compaction intervals in the separate September 20--23
    evaluation: 56/45 completed compactions and 51/41 within-session
    intervals without/with residual memory. Intervals do not cross restarts;
    two residual intervals exceed the display range but remain in the
    box statistics.}
    \label{fig:trading}
\end{figure}

\paragraph{Live Trading.}
Live Trading follows two concurrent Qwen paper-trading accounts under
the same context policy. At the September 23 cutoff, settled ROI is
$-11.86\%$ without residual and $+5.11\%$ with it; logged generation costs
are \$303.81 and \$170.30. ROI counts resolved markets against an initial
\$20,000 bankroll. These two trajectories differ in market selection and
timing, limiting return comparisons.
Both accounts retain the growth-and-reset pattern
(Figure~\ref{fig:trading}). Compactions fall from 56 to 45 alongside
fewer responses (5,336 versus 3,013). Per 1,000 responses, compactions
instead rise from 10.5 to 14.9: the lower total accompanies less interaction.
Mean live context decreases from 105,372 to 101,185 positions including
memory. Fewer calls and smaller contexts explain the lower generation
spend in this record.

\begin{figure}[t]
    \centering
    \begin{minipage}{0.28\linewidth}
        \centering
        \includegraphics[width=\linewidth]{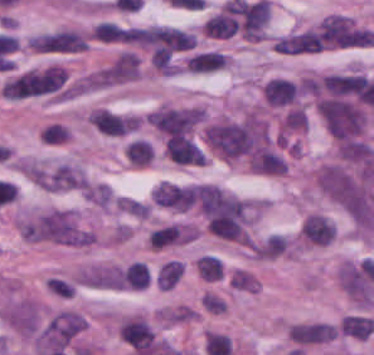}\\
        {\small Task image}
    \end{minipage}\hspace{0.06\linewidth}
    \begin{minipage}{0.28\linewidth}
        \centering
        \includegraphics[width=\linewidth]{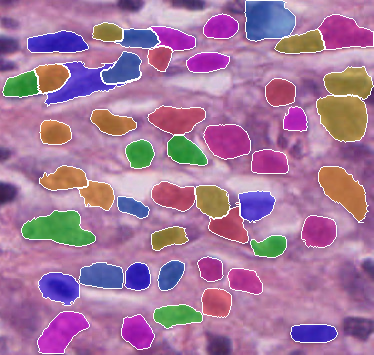}\\
        {\small Observed mask overlay}
    \end{minipage}
    \caption{Cell segmentation after repair. The task image and the exact
    overlay viewed by the actor are reproduced without pixel changes.}
    \label{fig:cells}
\end{figure}

\paragraph{Cell segmentation.}
In a Terminal-Bench cell-segmentation task, a refactor deletes helper
functions before compaction. The summary retains the known error and
backup identity but omits the exact edit commands. With summary and
memory available, the actor rereads both files, restores the helpers,
and passes all nine verifier tests (Figure~\ref{fig:cells}). The repair
illustrates complementary roles for summary, memory, and retrieval;
it does not establish which omitted details came from memory.

\section{Related Work}

Agent memory often combines summaries with retrieval. ReadAgent pairs
episode gists with original passages, while MemGPT separates working
and archival memory~\citep{lee2024readagent,packer2023memgpt}.
ReSum uses periodic summaries for long-horizon search; AgentFold learns
multi-scale context management within agent
trajectories~\citep{wu2025resum,ye2025agentfold}.
\remory{} keeps the textual checkpoint and retrieval interface, and learns
an additional continuous representation conditioned on the checkpoint.

AutoCompressors, Gist Tokens, and ICAE learn continuous representations
of longer text~\citep{chevalier2023autocompressors,mu2023gist,ge2023icae}.
LCLMs scale encoder--decoder compression and support selective expansion
of compressed segments~\citep{li2026lclm}.
Distribution matching provides a behavioral criterion for prompt
compression~\citep{wingate2022prompt}; recurrence and memory tokens
retain information across segments~\citep{dai2019transformerxl,bulatov2022rmt}.
Our objective applies this predictive criterion to the information gap
left by a summary, training the supplement through a frozen actor's
continuation predictions.

\section{Conclusion}
\label{sec:conclusion}

\remory{} learns a bounded memory supplement for contextual compaction.
The summary provides a readable account of prior work, soft tokens
carry additional predictive information, and retrieval supplies exact
records when needed. Reconstruction followed by summary-conditioned
continuation trains this interface while leaving the actor unchanged.
Improved source attribution, higher agent scores, and preferred next
actions support the usefulness of this residual formulation.
\label{sec:main-end}

\clearpage

\begingroup
\renewcommand{\bibfont}{\small}
\setlength{\bibsep}{4pt plus 1pt}
\bibliographystyle{iclr2027_conference}
\bibliography{arxiv/references}
\endgroup
\end{document}